\documentclass[letterpaper, 10pt, conference]{ieeeconf}  

\IEEEoverridecommandlockouts                              

\usepackage{amsmath}
\usepackage{amssymb}
\usepackage[hidelinks]{hyperref}
\usepackage{color}
\usepackage{graphicx}
\usepackage{subcaption}
\usepackage{multirow}
\usepackage{booktabs}
\usepackage{derivative}
\usepackage{mathtools}
\usepackage{algorithm}
\usepackage{algpseudocode}
\usepackage{times}
\usepackage{multicol}
\usepackage[dvipsnames]{xcolor}
\usepackage{flushend}
\usepackage{mdframed}
\usepackage[normalem]{ulem}

\DeclareMathOperator*{\argmax}{arg\,max}

\definecolor{navy}{rgb}{0.0, 0.0, 0.8}

\newcommand{\ph}[1]{{\textbf{#1}:}} 

\title{\LARGE \bf
SayComply: Grounding Field Robotic Tasks in Operational Compliance through Retrieval-Based Language Models
}

\author{\authorblockN{Muhammad Fadhil Ginting$^{1,2}$, Dong-Ki Kim$^{2}$, Sung-Kyun Kim$^{2}$, Bandi Jai Krishna$^{2}$, \\
Mykel J. Kochenderfer$^{1}$, Shayegan Omidshafiei$^{2}$, and Ali-akbar Agha-mohammadi$^{2}$}
\authorblockA{$^{1}$Stanford University, $^{2}$Field AI
}
}

\usepackage{flushend}

\usepackage{cleveref} 

\begin{document}
\maketitle


\begin{abstract}
This paper addresses the problem of task planning for robots that must comply with operational manuals in real-world settings. Task planning under these constraints is essential for enabling autonomous robot operation in domains that require adherence to domain-specific knowledge. 
Current methods for generating robot goals and plans rely on common sense knowledge encoded in large language models. However, these models lack grounding of robot plans to domain-specific knowledge and are not easily transferable between multiple sites or customers with different compliance needs. 
In this work, we present SayComply, which enables grounding robotic task planning with operational compliance using retrieval-based language models.
We design a hierarchical database of operational, environment, and robot embodiment manuals and procedures to enable efficient retrieval of the relevant context under the limited context length of the LLMs. 
We then design a task planner using a tree-based retrieval augmented generation (RAG) technique to generate robot tasks that follow user instructions while simultaneously complying with the domain knowledge in the database.
We demonstrate the benefits of our approach through simulations and hardware experiments in real-world scenarios that require precise context retrieval across various types of context, outperforming the standard RAG method.
Our approach bridges the gap in deploying robots that consistently adhere to operational protocols, offering a scalable and edge-deployable solution for ensuring compliance across varied and complex real-world environments. 
\\
Project website: {\color{navy} \href{https://saycomply.github.io}{saycomply.github.io}}.
\end{abstract}



    \section{Introduction}

Consider a mobile robot performing complex industrial inspections in unstructured environments,
such as performing maintenance checks that involve reading gauges, capturing thermal images, and inspecting various components~\cite{BostonDynamics2024,gehring2021anymal,agha2021nebula,shukla2016application}.
In industrial settings, these tasks are often supported by extensive documentation like operating manuals, training material, and additional written expert knowledge used to train human workers.
To complete these tasks, the robot must perceive the environment,  plan feasible actions, all while adhering to the established procedures and safety protocols outlined in these documents.
In this work, we address the challenge of task planning  for autonomous industrial robotics while ensuring the robot's compliance with relevant documentation and written expert guidelines.

\begin{figure}[t]
    \centering
    \includegraphics[width=0.48\textwidth]{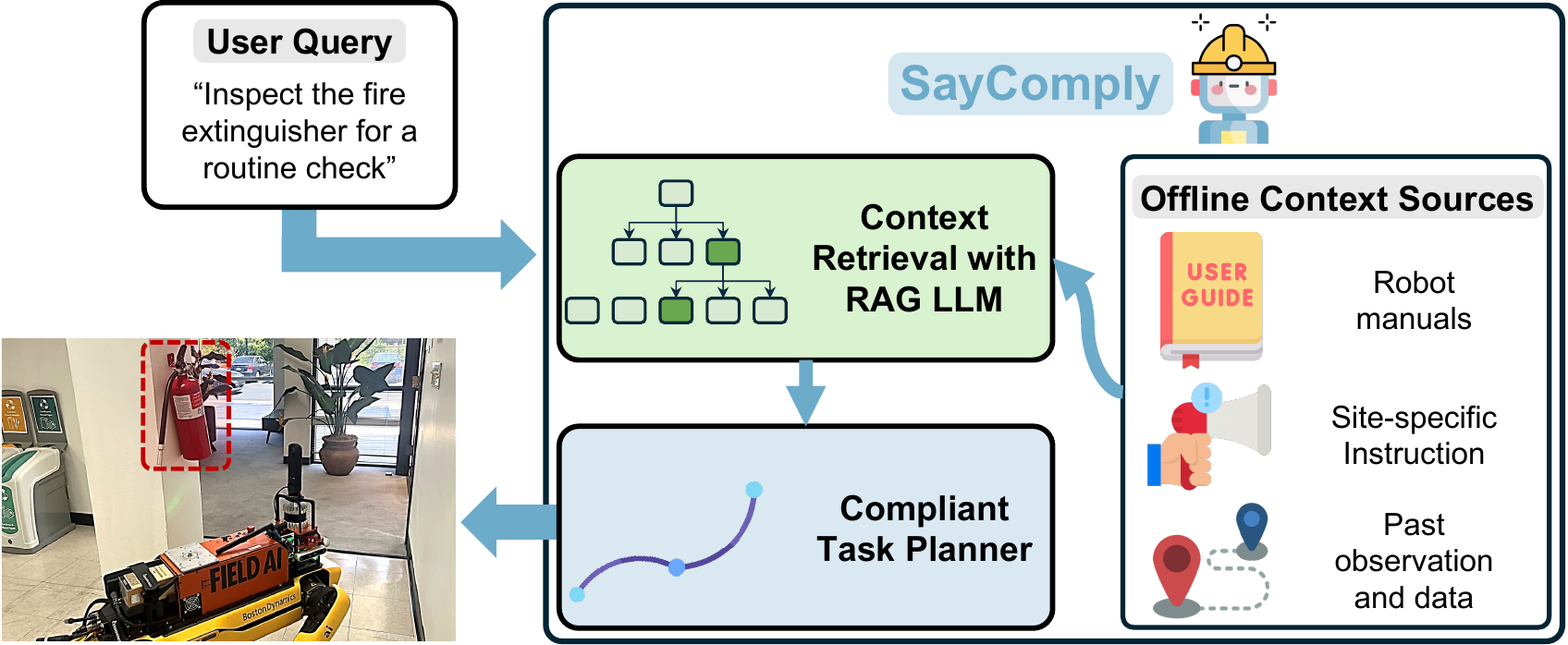}
    \caption{Autonomous robots operating in industrial settings need to \textit{comply} with operational procedures and past instructions from experts at the deployment sites while \textit{following} user instructions. SayComply grounds robot tasks with this information via a tree-based retrieval augmented generation (RAG), effectively leveraging data not commonly accessible to general-purpose LLMs.}
    \label{fig:cover_figure}
    \vspace{-0.6cm}
\end{figure}

Task planning using contextual information from expert knowledge and operational manuals presents two main challenges. 
The first challenge is \textit{grounding} robot plans to comply with operational information designed for humans rather than robotic AI. 
Robots differ from humans in embodiment, action spaces, and perception capabilities, leading to distinct ways of interacting with their environment. 
The second challenge is \textit{retrieving} relevant contexts from potentially extensive operational documents. 
On-edge robot reasoning models typically have limited context capacity, becoming inefficient when processing irrelevant information~\cite{yu2024defense}.
Effective context retrieval is crucial for compliant solution of tasks, underscoring the need for a new planning paradigm that grounds robot tasks in operational information.

To address these challenges, prior work suggests leveraging commonsense knowledge embedded in foundation models~\cite{brohan2023can, song2023llmplanner, liu2023llm+}. 
However, these approaches are limited to generic robotics problems, as field robotics in industrial settings often require proprietary knowledge not publicly available or usable for training general purpose foundation models in the first place. 
Another approach includes fine-tuning distinct models for each potential operational situation to ensure compliance, which is unfortunately infeasible for zero-shot deployment to new sites and customers. 
Foundation models like LLMs~\cite{brown2020language,devlin2019bert,openai2024gpt4technicalreport} and VLMs~\cite{radford2021learning,li2021align} have been shown to generate robot plans that conform with these sources of knowledge by conditioning them on the knowledge of the robot capability~\cite{brohan2023can} and environment observations~\cite{rana2023sayplan}. 
However, field robotics in industrial tasks often require reasoning over domain-specific and proprietary knowledge not available for training in general models. 
In this work, we investigate a new task generation approach that uses retrieval-based language models to integrate domain knowledge, enabling robotic task planning that complies with industry-specific information. 

In this paper, we propose SayComply, a novel method for robotic task planning that bridges the gap in deploying autonomous robots that comply with operational manuals and procedures in the field (\autoref{fig:cover_figure}). 
SayComply builds a hierarchical database of context sources, including documents, operational guidance instructions, detailed site-specific information, and robot embodiment safety and operating procedures. 
As LLMs have limited context length for in-context learning with such a database, we design a retrieval augmented generation (RAG) technique to retrieve relevant information given user instructions to ground robot task solution. 
Additionally, we develop compliant task planner using LLMs to generate plans, summarize information from the robot, and provide feedback to the user when instructions do not adhere to the compliance database. 
We validate our approach through experiments in different real-world scenarios in industrial inspections that require grounding to operational compliance, alongside extensive experiments in NVIDIA's high-fidelity Isaac simulator \cite{isaacsim} and on Boston Dynamics Spot robots.

In summary, our technical contributions are as follows:
\begin{enumerate}
    \item We present SayComply, a method leveraging retrieval-based language models to ground field robotic task planning with operational compliance;
    \item We design a hierarchical database of diverse operational context sources and develop a RAG technique to retrieve relevant information for task planning;
    \item We propose a method using LLMs to create compliant task plans and provide feedback to users on non-compliant instructions;
    \item We demonstrate SayComply's benefits in real-world scenarios requiring operational compliance through realistic simulations and hardware demonstration on a legged robot.
\end{enumerate}


\section{Related Work}
\label{sec:relworks}
\ph{Planning with Foundation Models}
Our work is related to approaches that use foundation models for robot planning. 
These approaches have been studied extensively~\cite{brohan2023can, song2023llmplanner, liu2023llm+, xie2023translating, izzo2024btgenbot, chen2022nlmapsaycan}
and applied to different problems in vision language navigation~\cite{shah2023navigation, zhou2023esc, yokoyama2024vlfm, BiggieMWH23, ginting2024seek}, 
manipulation~\cite{lin2023text2motion, huang2023voxposer}
embodied question answering, ~\cite{exploreeqa2024}
and anomaly detection~\cite{SinhaElhafsiEtAl2024}.
Particularly, our work is related to methods that enable robots to operate in large-scale unstructured environments by grounding its plan with 3D world representation~\cite{rana2023sayplan},
robot observation through few shot prompting ~\cite{song2023llmplanner}, 
demonstration videos~\cite{chiang2024mobility}, 
 and robot state summaries ~\cite{dlr205203}.
However, most of the works in current literature only use human commonsense knowledge encoded in general-purpose LLMs to generate plans and not using domain specific knowledge to ensure the robot plan is grounded with this information. 
Grounding to this knowledge robot is crucial, particularly for robots operating in industrial settings.
In this work, we design a task planning approach that is grounded to the domain-specific knowledge in addition to grounding with environment and robot embodiment information. 

\ph{Retrieval-augmented language model for robot planning} 
Retrieval-based approach for language model, such as augmented generation (RAG) ~\cite{NEURIPS2020_6b493230}, is a prominent framework for combining knowledge in a database with language generation~\cite{Borgeaud2021ImprovingLM, Gao2023RetrievalAugmentedGF, Ram2023InContextRL}. 
This enables LLMs to provide accurate and relevant information grounded in extensive external knowledge that is not contained in the LLM training datasets. 
In robotics, RAG techniques have been applied to retrieve relevant past user instructions and robot trajectory~\cite{xu2024prag, Kagaya2024RAPRP}, 
robot modules~\cite{dlr205203}, 
robot policy~\cite{Zhu2024RetrievalAugmentedEA}, 
and past driving scenarios~\cite{Yuan2024RAGDriverGD,Zhao2023ExpeLLA}. 
In this work, we design a new RAG technique to retrieve relevant knowledge of the environment, embodiment, and site operations, as well as to provide guidelines for LLM to generate plans that comply with this knowledge. 
We leverage recent techniques in tree-based RAGs~\cite{Sarthi2024raptor} to enable accurate and efficient retrieval, which cannot be achieved with naive RAG methods.

\ph{Statement of contribution} 
With respect to the current work in the literature, our work makes two key contributions. 
First, instead of relying on LLM commonsense knowledge and environment information to generate goals and plans for the robot, we ground robot plans with domain-specific knowledge so that the robot can operate in more critical settings that require compliance to certain manuals and procedures. 
Second, we demonstrate how to build and retrieve information from a database that combines data from robots with operational manuals and procedures, which are traditionally not designed to be consumed by robots.
To the best of our knowledge, SayComply is the first method to incorporate operational knowledge to ground field robotic task planning for real-world scenarios.

\begin{figure*}[t!]
  \vspace{0.25cm}
  \centering
  \includegraphics[width=0.8\textwidth,page=2]{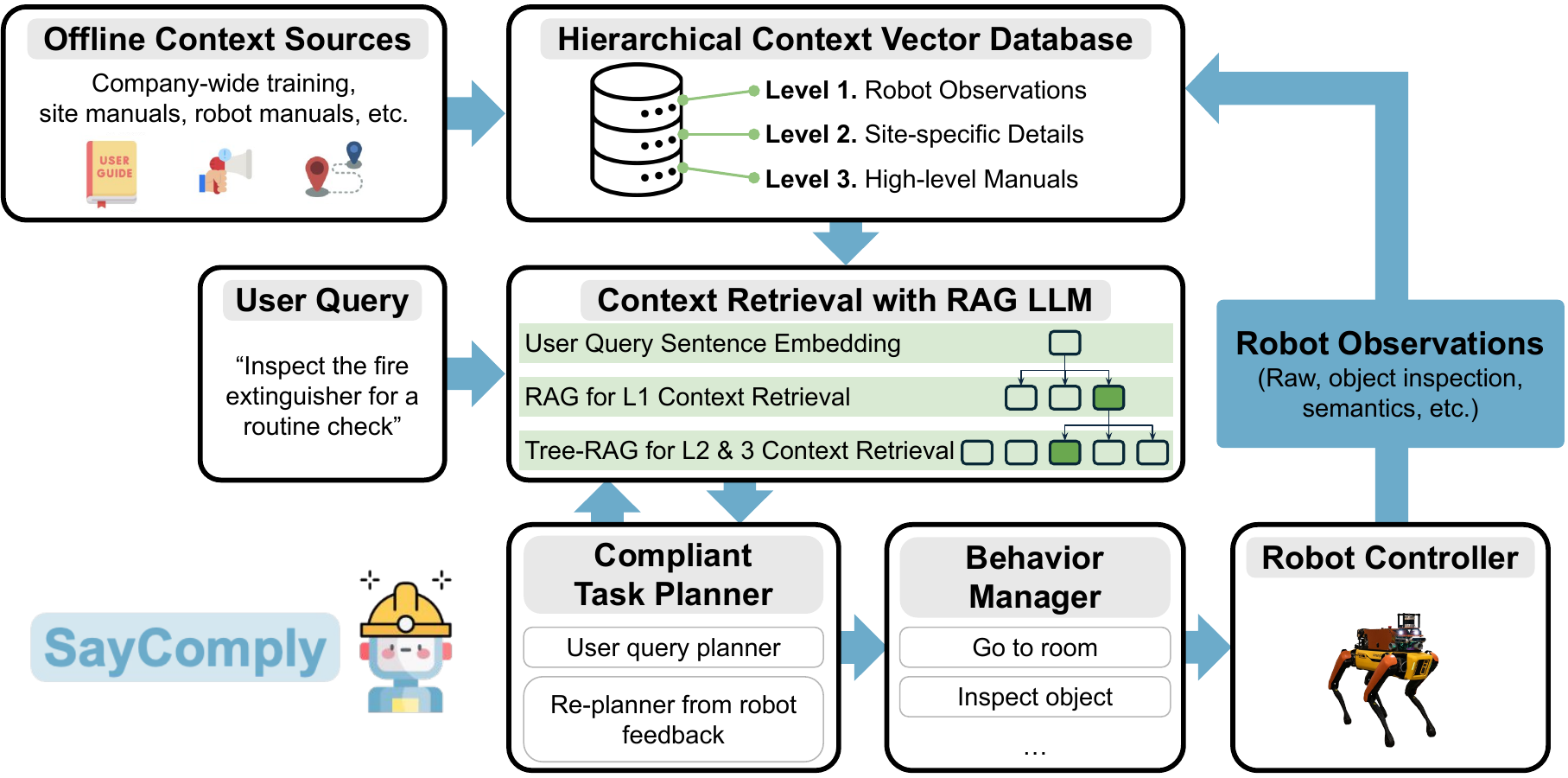}
    \caption{\textbf{SayComply system architecture.} Prior to robot deployment, we build a hierarchical context database from various written manuals and instructions. Next, given a user query, relevant context source is retrieved using a tree-based RAG and LLM method. Finally, the compliant task planner generates robot tasks based on the retrieved context. Tasks are executed by the robot through behavior manager and the robot observations are stored in the database.}
  \label{fig:sysarch}
  \vspace{-0.5cm}
\end{figure*}

\section{Compliant Task Planning using Retrieval-based Language Models}
\ph{Problem Statement}
Let $s=(x, G)$ denote the system state that consists of a robot state $x$ and a map of the environment $G$. 
The robot gets a query $q$ from a user expressed in natural language. 
The robot can perform different tasks $\pi \in \Pi$, and each task policy outputs a control action for the robot $\pi(s)=a$. To complete the user query $q$, 
the robot plans a sequence of tasks $\pi_{0:N}$, where $\pi_{N}$ is a task that responds back to the user on the completion of the query $q$. 
To fulfill the query, we select $\pi$ that maximizes $p(c_\pi \mid s, q)$, the probability of completing the query by executing $\pi$.
The robot can only estimate the query completion with probabilities because the robot does not have the full information about the state of the environment. 
Moreover, we want the robot to comply with the operational context provided in a database $D$ of instructions and manuals. 
Hence, the robot needs to also select a task that maximizes $p(y_\pi \mid s, q, D)$, the probability of being compliant with the operational context $D$ while executing $\pi$. 

\ph{Problem 1. Operation-Compliant Task Planning}
Given the robot and environment state $s$, a user query $q$, and a database of operational context $D$, compute a sequence of tasks $\pi_{0:N}$ that maximize the probability to complete the query and comply with the operational context: 
\begin{align}
    \pi_{0:N}^* = \argmax_{\pi_{0:N} \in \Pi} p(c_{\pi_{0:N}}, y_{\pi_{0:N}} \mid s, q, D).
\end{align}
%
This problem is highly challenging to solve. 
In order to ground robot tasks with operational compliance, a large amount of contexts in $D$ should be considered, but 
the language model that estimates the query completion and compliance has a limited context window. 
Moreover, providing irrelevant information to the model also increases the risk of not grounding with the correct context. 
Thus, we reformulate this problem with two sub-problems, operational context retrieval and operational context-based task planning.

\ph{Problem 2a. Operational Context Retrieval}
%
Given $s$, $q$, and $D$, select and retrieve a subset of the database $D^R$ that is relevant to the query $q$, 
where the retrieved database $D^R$ contains sufficient context to generate robot tasks that comply with the entire database $D$, and the size of $D^R$ is less or equal to a maximum context length $L$ of the language model:
\begin{align}
    &\text{find $D^R$} \\
    &\text{s.t. } p(y_\pi | s, q, D^R) \approx p(y_\pi | s, q, D), \\
    &\quad\,\, |D^R| \leq L.
\end{align}




\ph{Problem 2b. Operational Context-based Task Planning}
Given $s$, $q$, and $D^R$, compute a sequence of tasks $\pi_{0:N}$ that maximizes the multi-objective function of weighted query completion probability and compliance probability: 
\begin{align}
    \pi_{0:N}^*\!&=\!\argmax_{\pi_{0:N} \in \Pi} p(c_{\pi_{0:N}}\!\!\mid \!\! s, q, D^R)\!+\!\beta p(y_{\pi_{0:N}}\!\!\mid \!\! s, q, D^R),
\end{align} 
where $\beta \gg 1$ is a weight constant for the compliance probability. 
Note that $\beta \gg 1$ is to prioritize the compliance to safe robot operations on the field to the completion of the query, which may jeopardize the compliance.

\section{SayComply for Field Robotic Task Planning}
Having formulated the compliant task planning problem, this section introduces our method for field robotic task planning. At a high level, SayComply consists of the following steps: 
1) Prior to robot deployment, all the operational context sources $D$ is stored, summarized, and classified into a context source hierarchy using RAG techniques (\cref{sec:context_source_hierarchy}). 
2) When the robot is deployed and receives a user query $q$, relevant context sources are retrieved to ensure task planning compliance (\cref{sec:context_retrieval}). 
3) Finally, the planner composes the best sequence of tasks $\pi_{0:N}$ that completes the user query and sends the task to the robot for  execution. 
On every step of task execution, the planner generates an updated $\pi$ based on the new robot and environment state $s$.
\Cref{fig:sysarch} summarizes the architecture of SayComply.

\subsection{Context Source Hierarchy}\label{sec:context_source_hierarchy}
To retrieve relevant information efficiently, we build a hierarchy of context sources, represent every information as vector embeddings and store the information into a database. These vector embeddings are used for efficient search for semantic similarity during the retrieval process. 
This database is built offline prior to the robot deployment, and constructing it is computationally cheaper than fine-tuning an LLM.

All the relevant context sources are classified into three different levels (Context Levels 1-3) and three different context categories: \textbf{environment}, \textbf{operation}, and \textbf{embodiment}. 
Environmental context consists of information from written manuals associated with the site (e.g., floor plans or blueprints of the site). 
Operation context consists of written manual and verbal instructions of different procedures that can be performed by humans or robots at the site, such as inspection, surveillance, and maintenance tasks (e.g., oil and gas processing facilities inspection manual, office building maintenance manual). 
Embodiment context consists of information specific to the embodiment type and capabilities, such as movement, safety protocols, and operational constraints
(e.g., a legged robot operation manual~\cite{BostonDynamicsmanual}).

For each category, we propose to classify information into three different levels, where this hierarchy represents a structured context approach starting from the most immediate and specific to the most general and abstract context.
This structured approach enables effective retrieval from different context types and selection of relevant information given the user query. 

\textbf{Context Level 1: Current and past observations.} 
This context source consists of databases of past inspection data gathered by the robot or humans, current and history of robot's state and observation. 
Every inspection database, robot state, and observation are summarized in a sentence to help LLM select relevant data during context retrieval.

\textbf{Context Level 2: Site-specific details.} 
This context source consists of different instructions and guidelines to operate in a specific environment. 
This information comes from verbal instructions from a site expert or a brief summary coming from a longer context level 3 manual. 
Context level 2 can be seen as basic information and instructions that is given by a site expert to a new human operator and robots before being deployed at the site. 


\textbf{Context Level 3: High-level manuals.} 
This context source consists of different documents that are used as references for human operators to ensure operational compliance. 
Example of context level 3 includes inspection manuals, safety document, environment blueprints, and robot operating and safety manual. 
Every document in this level is referenced by at least one instruction from context level 2 to help with referencing correct information during retrieval. 

\subsection{Context Retrieval with RAG LLM}\label{sec:context_retrieval}
Context retrieval is crucial for SayComply to ground task planning with operational compliance. 
Given human instruction, it selects relevant information from the context source database and passes the information to the task planner. 
The LLM-based task planner is constrained by the limited context length and attention of the LLM, so only a subset of the database under a specified maximum context length is retrieved.
We employ two different retrieval techniques. 

\ph{Context level 2 and 3 retrieval}
To retrieve relevant instructions and manuals from context levels 2 and 3, we design a tree-organized retrieval method. 
Given the user query, we encode the query as a vector embedding and retrieve top 2 most similar entry in level 2 context sources using cosine similarity. 
Then, we retrieve the manual most similar to the user query among the level 3 context sources, which are pointed by the retrieved level 2 context sources.

\ph{Context level 1 retrieval}
We query an LLM to select relevant data from context level 1. 
We find cosine similarity is less effective to select historical and numerical data with less semantic information. 
We provide a list of title name of the database in context level 1 with a short description and query LLM to select the database that can be useful to complete the query.


\subsection{Compliant Task Planner}\label{sec:task_planner}
Given the user query and the context retrieved from the database, the LLM-based task planner generates a sequence of robot tasks and responds to the user. 
The retrieved context is used to constrain the robot task space and provide all the context needed to generate plans. The planner generates the best robot task that completes the user queries.  

The planner prompt is provided with different type of robot tasks that can be recognized and executed by the robot. 
While the task types can be set based on robot capability and applications, for industrial inspection scenarios the most common classes of tasks include: go-to object or rooms, search for an object, inspect (e.g., read, scan or measure) an object or a room, and so on. 
The prompt also includes an option to respond directly to the user when the user query violates operational compliance. 
We design a single prompt to generate plans in two different cases: task planning from a new user query, and task replanning from robot feedbacks.

\textbf{Task planning from a user query} generates a sequence of tasks given a user query. 
The LLM is prompted to check whether the query adheres to retrieved contexts, and is queried to output the best sequence of plans for the robot that has a high-level of compliance and completes the query. 
It is also prompted to provide a justification of choosing the plans according to the retrieved contexts. 
Only the first task is sent to the robot behavior manager for execution. 
While querying a sequence of tasks can guide the LLM to generate a complete task plan, 
executing only the first task allows the planner to replan based on new observations and query completion status.
Subsequent plans are cached with the original user query for replanning. 

\textbf{Task replanning from robot feedback} is triggered after the robot finishes executing a task and reports its completion status. 
In addition to querying LLM with the same context as the first planning from user query, we append the status of the previously-executed task and the cached sequence of next tasks. 
The replanner outputs an updated task sequence (either the same plan or a revised plan) and sends a new task to the robot. 
Replanning is performed until all tasks to fulfill the query are attempted. 
Finally, the planner composes an answer for the user on the completion status. 

\begin{figure*}[t!]
  \centering
\includegraphics[width=1.0\textwidth]{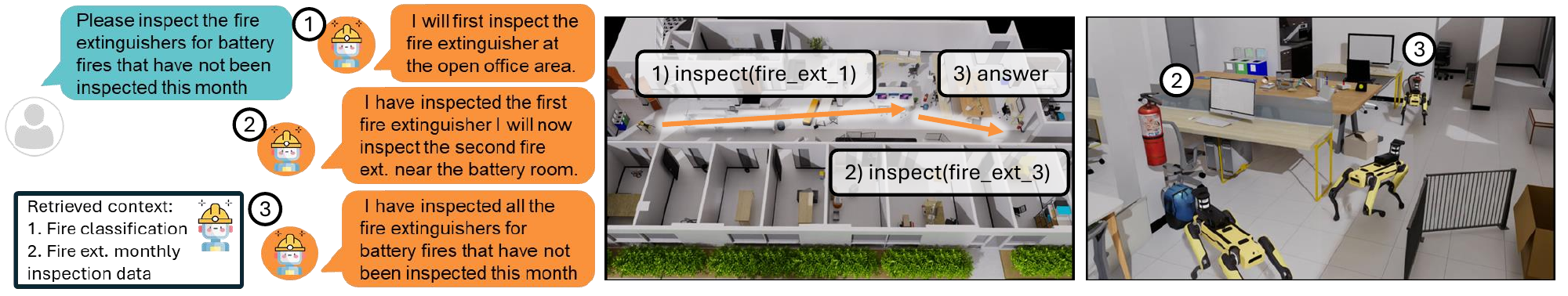}
    \caption{\textbf{Simulation results.} The left panel shows the user query and robot answers while executing the plans. The middle and right panels show the robot task execution inspecting the fire extinguishers.}
  \label{fig:sim_results}
  \vspace{-0.45cm}
\end{figure*}

\section{Experimental Results}
We evaluate the performance of our approach in real-world scenarios requiring operational compliance, both in simulations and on a legged robot platform. 

\subsection{Scenario description}
Given that no existing benchmarks exist for field robotic task planning under operational compliance, we design a new set of experiments based on real-world use cases.

\ph{Real world scenarios} 
We present three distinct use cases wherein a mobile robot is assisting humans in real-world operations and inspections:

\textbf{1) Industrial inspection in oil \& gas and manufacturing} is a critical use cases where robots can help ensure continuous operation at plants, by reading gauges, capturing thermal images, and reporting anomalies for predictive maintenance.
    
\textbf{2) Office operations and maintenance} require robots to assist humans in tasks including regular maintenance checks (e.g., inspecting air conditioning systems), guiding personnel through safety and emergency procedures, and supporting daily office activities like providing office orientation, locating meeting rooms, and finding items. 

\textbf{3) Embodiment-aware operation in the field} is related to robot task planning that needs to adhere to robot operation and safety manuals. The manual includes safety information when operating near humans, navigating challenging terrains, adjusting locomotion parameters, and providing guidelines on navigation and environmental conditions.

We compile all relevant instructions and manuals and simulate past observations for these three use cases, building a hierarchical database of context sources that includes 62 different manuals, instructions, and files.

\ph{User queries} Based on real-world scenarios, we generate 70 different user queries for our experiments. These represent a balanced mix of queries requiring compliance with various levels of context sources. 
We also include non-compliant user queries to evaluate how effectively our approach rejects inappropriate requests.

\ph{Environment description} 
All simulations and hardware experiments are conducted within the Field AI office environment, using a high-fidelity digital twin of the space to perform extensive and parallel tests before proceeding with hardware experiments.

\begin{figure}[t]
    \centering
    \includegraphics[width=0.48\textwidth]{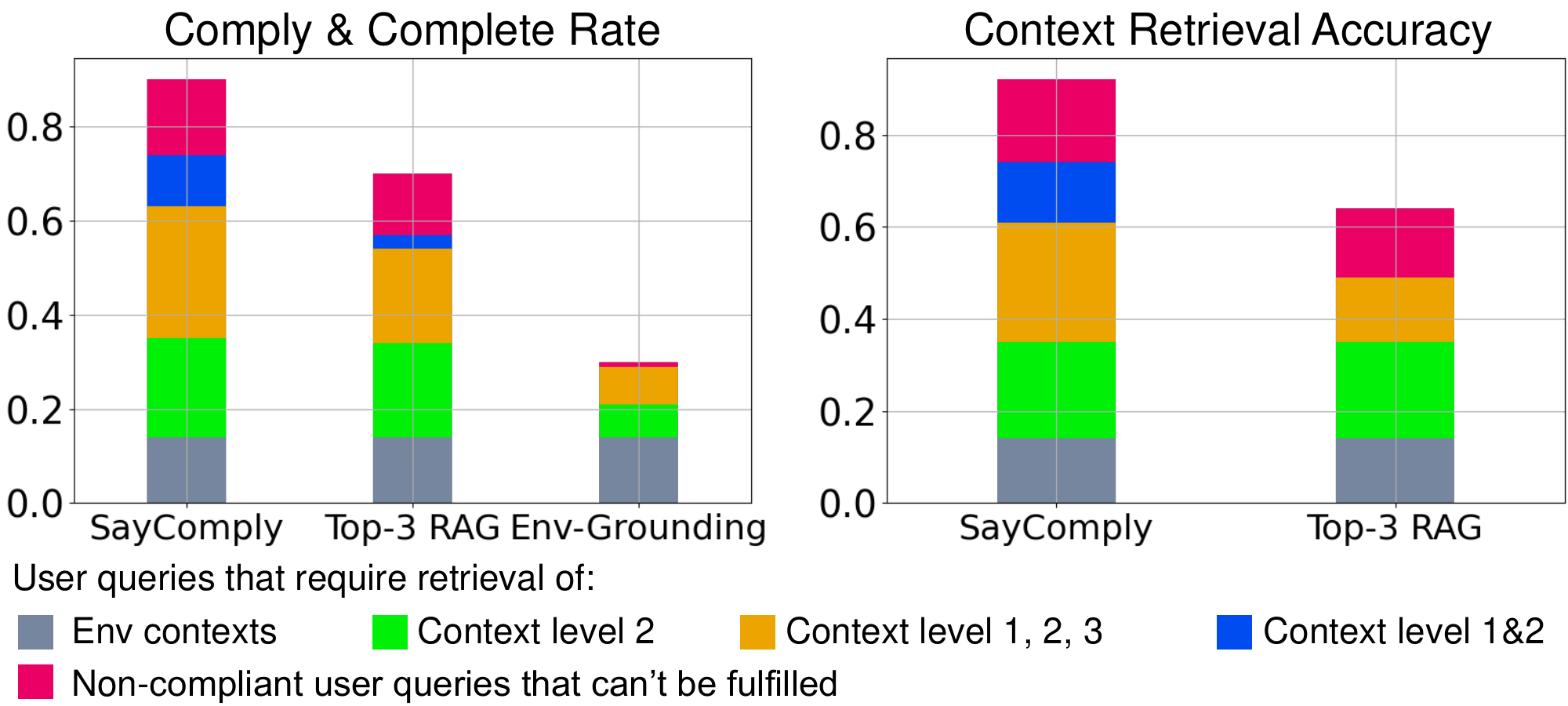}
    \caption{The compliance \& completion rate, and context retrieval accuracy from different types of user queries.
    }
    \label{fig:sim_querytypes}
    \vspace{-10pt}
\end{figure}

\begin{table}[t!]
\renewcommand{\arraystretch}{1.27}
\caption{Simulation results showing the percentage of user queries each method successfully addressed, including the ability to \textit{comply}, to \textit{comply \& complete}, and to retrieve the context correctly.}

\centering
\begin{tabular}{@{}lcccc@{}}
\toprule
\textbf{Method} & \textbf{Comply} & \textbf{Comply\&Complete} & \textbf{Context Retrieval} \\
\midrule
\textbf{Env-Grounding} & 32.9\%  & 30.0\% & N/A \\
\textbf{Top-3 RAG}  & 72.9 \% & 70.0\% & 65.7\% \\
\textbf{SayComply} & \textbf{91.4\%} & \textbf{91.4\%} & \textbf{92.9\%} \\
\bottomrule
\end{tabular}
\label{tab:method_results}
\vspace{-4pt}
\end{table}

\subsection{Simulation results}
We evaluate our approach in simulations in Issac Sim \cite{isaacsim} in the office environment. 
The user queries are inputted through a web interface, and the generated robot task is executed in the simulator by a behavior manager. 
In all the experiments, we use Open AI's GPT 4 model~\cite{achiam2023gpt}. 
\Cref{fig:sim_results} shows the simulation environment and how our approach generates and executes the robot tasks.


We evaluate SayComply against two approaches: 
\begin{enumerate}
    \item \textbf{Env-Grounding}. This approach only uses environmental context, including 3D scene graphs and building floor plans, to generate robot tasks without retrieving operational and embodiment contexts.
    \item \textbf{Top-3 RAG}. This approach retrieves the 3 most relevant context sources to the user query, and is used to evaluate the benefits of our context retrieval approach.
\end{enumerate}

\begin{figure*}[t!]
  \centering
\includegraphics[width=1.0\textwidth]{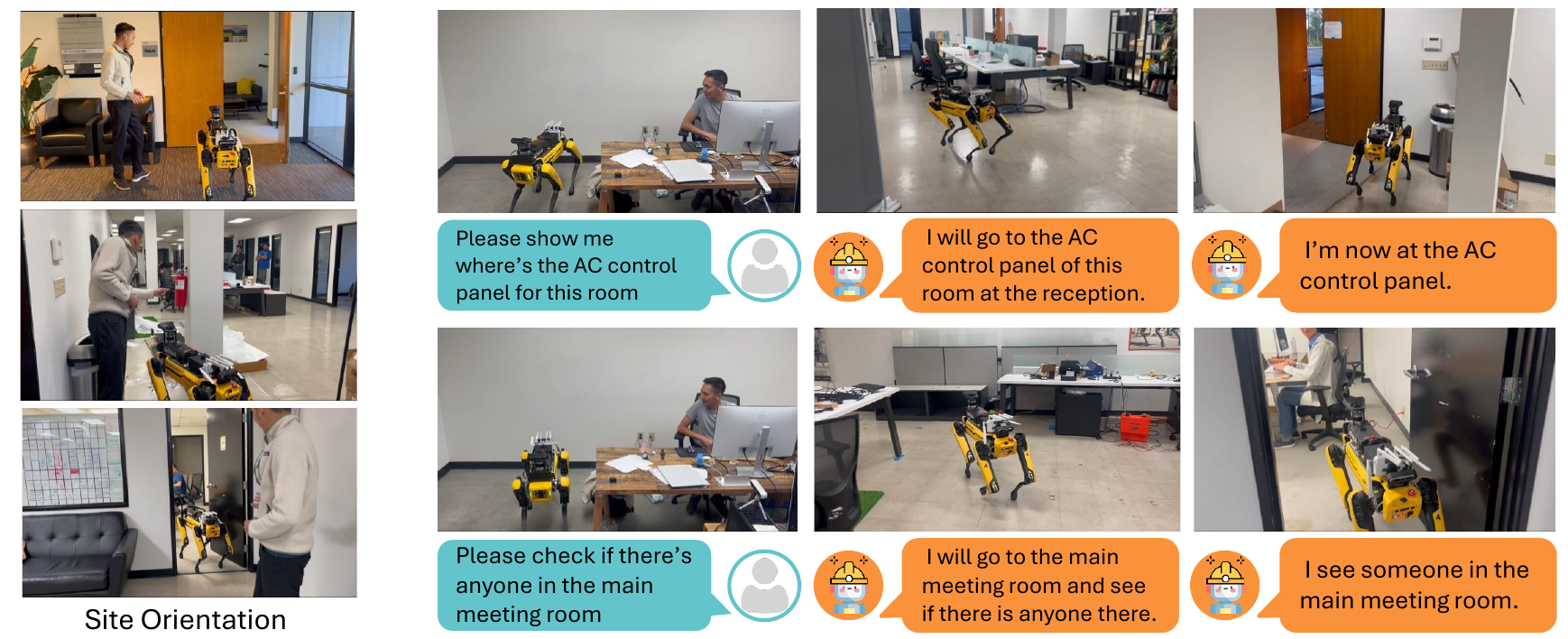}
    \caption{\textbf{Experiment results on hardware.} The left panel illustrates an expert user first providing site orientation to the robot. The right panel shows the subsequent robot task execution given the user queries.}
  \label{fig:hw_results}
  \vspace{-0.45cm}
\end{figure*}

\ph{Evaluation metrics} We use the following metrics to evaluate the methods: 1) \textbf{Comply}: Percentage of user queries for which the method generates plans that comply with the context database $D$, 2) \textbf{Comply \& Complete} Percentage of user queries for which the method generates plans that both comply with $D$ and fully address the user query, 3) \textbf{Context Retrieval}: Percentage of user queries for which the method successfully retrieves context sources with enough information to comply with $D$ and complete the query.

\autoref{tab:method_results} compares the performance of all methods, and \autoref{fig:sim_querytypes} examines the performance by user query category. 
\textit{Env-Grounding} method is the least performant, as it only fulfills queries requiring environmental information and general operational context that is already available in the LLM (e.g., information on different categories of fire extinguishers). 
Our approach performs the best in all of the metrics. 
We find that correct context sources is crucial and strongly correlated with the other metrics. 
We observe the reason the \textit{Comply \& Complete}, and \textit{Comply} rates are lower to the context retrieval is due to the LLM misinterpreting the context correctly. 
For example, we find that the LLM often misinterprets tabular information in past observation data and generates non-compliant or non-complete robot tasks. 
For \textit{Top-3 RAG}, we observe the \textit{Comply \& Complete} rate exceeds the \textit{Context Retrieval} because the LLM can generate correct tasks even when the retrieved information is incomplete (e.g., only retrieving two out of three required contexts).

We observe a significant difference in the retrieval accuracy between SayComply and \textit{Top-3 RAG}.
While both methods perform well when retrieving queries requiring level 2 context, we notice a difference in retrieval performance when retrieving queries that require level 1 and 3 context. 
In contrast to shorter user instructions explicitly designed for humans or robots, retrieving level 1 and 3 context is more challenging. 
Level 1 context mainly consists of tabular data and report logs, which are difficult to compare using sentence embeddings in standard RAG retrieval. 
Meanwhile, retrieving level 3 context is more challenging as the information needed can be in different parts of the document. 
Using tree-RAG enables narrowing down the search based on the context level 2 retrieval.
The results highlight the benefit of our tree-based RAG and LLM-based context level 1 retrieval for operational-compliant robotic task planning problems.


\subsection{Hardware Results}

We demonstrate the efficacy of our approach on hardware by deploying SayComply on the Boston Dynamics Spot legged robot, which is equipped with a LIDAR and a camera for navigation and semantic observation~\cite{ginting2024semantic, bouman2020autonomous}.
A laptop, connected to the internet with a web-based user interface, is used for inputting user queries, sending tasks, and remote monitoring.
The laptop hosts the context database, retrieval, and task generation pipelines. 
We experiment with 15 different user queries in office operations and maintenance.

Prior to the experiments, we verbally give the robot site orientation while it autonomously walks around the office.
Received instructions are converted to text and associated with specific rooms where they are received.
This procedure mirrors site orientations typically conducted in industrial settings, further highlighting the practicality of our approach.

In the experiments, SayComply enables the robot to execute various user queries~(\autoref{fig:hw_results}). The robot completes the queries requiring compliance and an understanding of different operational contexts for real-world inspection and maintenance. 
Users can remotely send the query, monitor the robot, and receive real-time updates from the robot via the laptop interface. 
SayComply's efficient context retrieval enables compliant task planning without reliance on a long-context-window that can not be hosted on the robot or when the robot is deployed in sites without reliable internet access. 
These experiments demonstrate the practicality of our approach for real-world robot operation requiring operational compliance across diverse use cases.


\section{Conclusion} 

We have presented SayComply, a method for field robotic task planning grounded in operational compliance. 
SayComply retrieves relevant contexts from a database of operational, environment, and robot embodiment manuals and procedures to generate robot tasks. 
Our retrieval and LLM-based task generation method enables robots to plan tasks complying with domain-specific knowledge. 
Our simulations and hardware experiments demonstrate the efficacy of our approach in various real-world industrial inspection tasks. 
This work marks an important step towards enabling autonomous task planning for robot operation in various use cases that require compliance with domain-specific protocols. 
For future work, we plan to design a role-based task planner that allows or prohibits certain robot tasks based on the level of authority of the person that instructs the robot.

\vspace{-5pt}

\bibliographystyle{IEEEtran}
\bibliography{references}

\end{document}